%% file: FAccT_2022_final.tex
\documentclass[screen]{acmart}

\AtBeginDocument{%
  \providecommand\BibTeX{{%
    \normalfont B\kern-0.5em{\scshape i\kern-0.25em b}\kern-0.8em\TeX}}}

\setcopyright{acmcopyright}
\copyrightyear{2022}
\acmYear{2022}
\acmDOI{}
\acmISBN{}

\acmConference[FAccT '22]{FAccT '22: Conference on Fairness, Accountability and Transparency (ACM FAccT)}{June 21--24, 2022}{Seoul, South Korea}
\acmBooktitle{FAccT '22: ACM Conference on Fairness, Accountability and Transparency, June 21--24, 2022, Seoul, South Korea}

\usepackage{algorithm}
\usepackage{algcompatible}
\usepackage{amsthm}
\usepackage{caption}
\usepackage{subcaption}

\makeatletter
\newcommand\nocaption{%
    \renewcommand\p@subfigure{}
    \renewcommand\thesubfigure{\thefigure\alph{subfigure}}
}

\newenvironment{proofof}[1]{\par
  \pushQED{\qed}%
  \normalfont \topsep6\p@\@plus6\p@\relax
  \trivlist
  \item[\hskip\labelsep
        \bfseries
    Proof of #1\@addpunct{.}]\ignorespaces
}{%
  \popQED\endtrivlist\@endpefalse
}
\makeatother

\begin{document}

\newcommand{\Prob}{\mathbb{P}}
\newcommand{\EOpp}{\textit{EOpp }}
\newcommand{\EOdds}{\textit{EOdds }}
\newcommand{\bern}{\textit{Bernoulli }}
\newcommand{\unif}{\textit{Uniform }}
\newcommand{\Exp}[1]{{\mathbb E}\left( #1 \right)}
\newtheorem{assumption}{Assumption}

\title{Achieving Fairness via Post-Processing in Web-Scale Recommender Systems}
\titlenote{None of the authors received any third-party funding for this work.}

\author{Preetam Nandy}
\email{pnandy@linkedin.com}
\affiliation{%
  \institution{LinkedIn Corp.}
  \city{Sunnyvale}
  \country{U.S.A.}
}

\author{Cyrus DiCiccio}
\authornote{These authors contributed to this work while they were at LinkedIn Corp.}
\affiliation{%
  \institution{while at LinkedIn Corp.}
  \city{Sunnyvale}
  \country{U.S.A.}
}
\email{cjd48@cornell.edu}

\author{Divya Venugopalan}
\email{dvenugopalan@linkedin.com}
\affiliation{%
  \institution{LinkedIn Corp.}
  \city{Sunnyvale}
  \country{U.S.A.}
}

\author{Heloise Logan}
\email{hlogan@linkedin.com}
\affiliation{%
  \institution{LinkedIn Corp.}
  \city{Sunnyvale}
  \country{U.S.A.}
}

\author{Kinjal Basu}
\email{kbasu@linkedin.com}
\affiliation{%
  \institution{LinkedIn Corp.}
  \city{Sunnyvale}
  \country{U.S.A.}
}

\author{Noureddine El Karoui}
\authornotemark[1]
\affiliation{%
  \institution{while at LinkedIn Corp.}
  \city{Sunnyvale}
  \country{U.S.A.}
}
\email{nkarouiprof@gmail.com}

\renewcommand{\shortauthors}{P.\ Nandy, C.\ DiCiccio, D.\ Venugopalan, H.\ Logan, K.\ Basu and  N.\ El Karoui}

\begin{abstract}
  Building fair recommender systems is a challenging and crucial area of study due to its immense impact on society. We extended the definitions of two commonly accepted notions of fairness to recommender systems, namely equality of opportunity and equalized odds. These fairness measures ensure that equally ``qualified'' (or ``unqualified'') candidates are treated equally regardless of their protected attribute status (such as gender or race). We propose scalable methods for achieving equality of opportunity and equalized odds in rankings in the presence of position bias, which commonly plagues data generated from recommender systems. Our algorithms are model agnostic in the sense that they depend only on the final scores provided by a model, making them easily applicable to virtually all web-scale recommender systems. We conduct extensive simulations as well as real-world experiments to show the efficacy of our approach.
\end{abstract}

\keywords{}

\maketitle

\input{texfiles/introduction}

\input{texfiles/relatedWork}

\input{texfiles/cdftransfo}
\input{texfiles/equalOdds}

\input{texfiles/positionBias}

\input{texfiles/experiments}
\input{texfiles/application}
\input{texfiles/conclusion}

\bibliographystyle{ACM-Reference-Format}
\bibliography{biblioFairness}

\input{texfiles/appendix}

\end{document}

%% file: texfiles/introduction.tex
\section{Introduction}
\label{sec:intro}
Fairness in classification \cite{agarwal2018reductions, bechavod2017penalizing} and ranking problems \cite{Geyik19} has been an active area of research in recent years due to its tremendous influence on society as a whole \cite{barocas2017fairness}. As more and more businesses rely on machine learning (ML) algorithms to recommend goods and services, treating affected individuals fairly is becoming ever more important. ML-based systems may contain implicit biases and can serve to reproduce or reinforce the biases present in society \cite{buolamwini2018gender, zou2018ai}. It is crucial to be able to mitigate these biases, especially in large-scale industrial applications. 

Fairness mitigation strategies commonly fall into one of three categories: pre-, in-, or post-processing. Pre-processing modifies training data to reduce potential sources of bias, often by removing features that are correlated with protected attributes
(\cite{zemel13, flavio17, gordaliza19}).  In-processing (also known as training-time) mitigation methods modify the model training objective to incorporate fairness, often by adding constraints or regularization penalties
\cite{KamishimaPenalized2012,bechavod2017penalizing,mary2019fairness,pmlr-v54-zafar17a,agarwal2018reductions}. Finally, post-processing methods transform model scores to ensure fairness according to a provided definition. Post-processing methods learn (protected-attribute-specific) transformations of model scores to achieve fairness objectives \cite{Hardtetal_NIPS2016, pleiss2017fairness, kamiran2012decision}. They are very appealing in industrial practice as they do not require changes to an existing model training pipeline. Virtually any model can be easily adjusted by a post-processing algorithm to achieve the desired fairness goal.

In this paper, we derive post-processing approaches providing fairness for ranked lists of items generated by a recommender system.
Hardt et al. \cite{Hardtetal_NIPS2016} introduce post-processing methods for equality of opportunity and equalized odds in the binary classification setting. However, these methods are not directly applicable to the ranking problems where fairness needs to hold with respect to the ranks of the items. Ranking problems are further complicated by the presence of position bias \cite{JoachimsUnbiasedLearningToRankFromBiasedFeedback, positionBiasEstimationGoogle2018}, i.e., the bias in an end user's response depending on an item's position, which is not a concern in the binary classification setting. Geyik et al. \cite{Geyik19} provides an algorithm for fair ranking (informed by an approximation of equality of opportunity) by modifying the classification definition. This work does not address position bias, and the methodology does not extended to other notions of fairness, such as equalized odds. 

Recently, there has been extensive work focused on framing fairness in ranking as an optimization problem maximizing relevance subject to fairness constraints \cite{celis2017ranking, joachims2018, SinghJoachims2019}. These procedures are also limited in the types of fairness they can accommodate (largely addressing variations of equal exposure) and cannot be directly applied to fairness definitions incorporating outcomes such as equality of opportunity or equalized odds \cite{Hardtetal_NIPS2016}. They are also impractical for many internet applications, where latency concerns may pose problems for solving an optimization problem for real-time queries. 

We formally extend the definitions of equality of opportunity and equalized odds from the binary classification \cite{Hardtetal_NIPS2016} to the ranking context, provide a causal interpretation of the equalized odds condition (Section \ref{subsection: causal interpretation}), and provide scalable post-processing techniques for mitigation of bias identified through these definitions, which has not been addressed by previous literature on fairness in ranking. We also explicitly handle the position bias issue and suggest simple mechanisms for controlling the fairness versus performance trade-off. Our novel method of enforcing equalized odds in ranking is highly flexible, and we show how it can be easily extended to tackle multiple outcomes and appropriate relaxations. We perform extensive experiments to showcase the efficacy of our procedure by considering simulated, public, and real-world datasets.


The remainder of the paper is organized as follows. We first define and then develop post-processing techniques to adjust for equality of opportunity and equalized odds in rankings (Section \ref{section:equalityOfOpportunity} and Section \ref{sec:EOdds} respectively). We address the position bias issues present in the ranking context for both the mechanisms in Section \ref{sec:positionbias}. We empirically show the efficacy of our approach in Section \ref{sec:empirical} before concluding with a discussion in Section \ref{sec:discussion}. Proofs of all results are given in the Appendix in the supplementary material. We end this section with a brief overview of the related literature.

%% file: texfiles/relatedWork.tex
 
\subsection{Related Work}
\label{sec:related}

Most early work in fairness has been on classification tasks and strives to achieve fairness notions such as equalized odds on protected attributes \cite{agarwal2018reductions, Hardtetal_NIPS2016,  feldman2015certifying, ZafarEtAl17, zafar2017bfairness, goh2016satisfying, wu2019convexity}. Also, many techniques for training models that guarantee other definitions of fairness, such as equality opportunity \cite{Hardtetal_NIPS2016} and demographic parity, have been widely studied in recent years \cite{dwork2012fairness, pmlr-v28-zemel13, goel2018non, johndrow2019algorithm}. Penalty methods for incorporating fairness constraints during model training has received a great deal of attention \cite{KamishimaPenalized2012, bechavod2017penalizing, mary2019fairness, pmlr-v54-zafar17a}. Although they are sound theoretical methods, many of them do not scale well due to the iterative nature of these algorithms \cite{agarwal2018reductions} and hence can become a bottleneck in many large-scale applications. Thus, post-processing techniques, which are model-agnostic, are often preferable. 

Many large-scale recommender systems are ranking systems, returning a list of ranked results to the users. Although there are several scalable post-processing methodologies \cite{Hardtetal_NIPS2016, pleiss2017fairness, kamiran2012decision} for fairness, most of them focus on classification problems, and they do not directly translate to the ranking problem. Moreover, position bias plays a critical role in these ranking frameworks, which we address through our post-processing reranker. Several works handle fairness in recommender systems by solving a constrained optimization to optimize for relevance, subject to fairness constraints \cite{joachims2018,celis2017ranking}. These techniques are broadly applicable but suffer from two drawbacks. First, they are limited in the types of fairness they can address and are limited to addressing variations of equal exposure (and cannot be applied to fairness definitions incorporating outcomes such as equality of opportunity). Second, they are impractical for many internet applications, where latency concerns may pose problems for solving an optimization problem for real-time queries. \cite{Geyik19} provides an algorithm for fair recommendations informed by an approximation of equality of opportunity through an approximation to the classification definition. However, this work lacks a meaningful extension of this fairness notion to the ranking setting and does not address equalized odds. There have also been works addressing fairness in Learning-to-Rank applications \cite{SinghJoachims2019, morik21} which is beyond the scope of this paper.


%% file: texfiles/cdftransfo.tex
%
%

\section{Equality of Opportunity in Rankings}\label{section:equalityOfOpportunity}

Throughout, we assume that a machine learning model predicts an outcome $Y$ from available features $X$ resulting in a score $s(X)$ (or simply $s$ for compactness). These predictions are used to provide ranked recommendations of people or items where each recommendation is assumed to have a categorical group characteristic $C$ (e.g., protected attribute status). Before discussing equality of opportunity for ranking, we will recall the definition of equality of opportunity (\textit{EOpp}) for binary classification \cite{Hardtetal_NIPS2016}. In the binary classification context, machine learning models yield a binary prediction $\hat{Y}$ which can but need not be derived from a score $s(X)$.

\begin{definition}[\EOpp in classification]\label{definition: EOpp} A binary predictor $\hat{Y}$ satisfies equal opportunity with respect to a (protected) characteristic $C$ if $\hat{Y}$ is independent of $C$ given $Y=1$, that is, $$\Prob(\hat{Y}=1 \mid C=c_1,Y=1) = \Prob(\hat{Y}=1 \mid C=c_2,Y=1),~ \text{for all $c_1, c_2$.}
$$ In other words, $\hat{Y}$ is independent of $C$ given that $Y=1$.
\end{definition}

A useful extension of this condition beyond binary model outputs is to the case when the model outputs scores. 
 In this setting, we require that the distribution of model scores be be independent of $C$ given that $Y=1$ (see, e.g., \cite{mary2019fairness}, Section 2.1). For threshold based classifiers (i.e., ones for which \ $\hat{Y}(X)=1$ if and only if $s(X)>t$ for an appropriate threshold $t$), this would ensure \EOpp for the associated classifiers at all thresholds.

\begin{definition}[\EOpp in rankings]\label{definition: ranking EOpp} A scoring function satisfies equal opportunity with respect to a (protected) characteristic $C$ and score $s(X)$ if for all $c_1$, $c_2$ and $t$,
$$
\Prob(s(X) \leq t \mid C=c_1,Y=1) = \Prob(s(X) \leq t \mid C=c_2,Y=1),~ \text{}.$$ 
\end{definition}

In industry applications, Definition \ref{definition: ranking EOpp} is a natural requirement as scores could be passed downstream to other machine learning systems before yielding final recommendations. Even in the case of classification, applying the ``ranking'' definition of \EOpp can be preferable. One reason is that in many internet applications, the threshold $t$ for a score based classifier is chosen through A/B experimentation to yield a desirable tradeoff of business metrics \cite{Agarwal:oms}. As such, it is generally not known a priori which threshold is needed, and this definition affords the robustness and flexibility of guaranteeing fairness regardless of the threshold chosen. 
Note that the solution proposed in \cite{Hardtetal_NIPS2016} is based on Definition \ref{definition: EOpp}, while in the following subsection, we propose a solution based on Definition \ref{definition: ranking EOpp}. 

\subsection{Algorithm to achieve Equality of Opportunity: }
In the following lemma, we present a simple post-processing algorithm that achieves \EOpp at all thresholds $t$ simultaneously. Define the cumulative distribution function (CDF) of scores in group $C = c$ and $Y=1$ by
\[
F_{c,1}(t) = \Prob(s(X) \leq t \mid C=c_1,Y=1) = \Prob(s(X) \leq t \mid C=c,Y=1)~.
\]
Applying an appropriate CDF transformation to the model scores achieves \EOpp.

\begin{lemma}[Algorithm for \EOpp]\label{lemma: CDF transformation}
Let $F_{c,1}(\cdot)$ be the cumulative distribution function (CDF) of scores in group $C = c$ and $Y=1$.  Then for each $c$, the transformation of the scores of group $C=c$ as $s(X) \to F_{c, 1}(s(X))$, will guarantee \EOpp for all thresholds $t$. 
\end{lemma}

The CDF transformations in Lemma \ref{lemma: CDF transformation} maps the scores into $[0, 1]$. We can apply an additional transformation $F^{-1}(\tilde{F}(\cdot))$ to bring the scores back to the original scale, where $F$ and $\tilde{F}$ are the CDF of the scores before applying any transformation and after applying the transformations in Lemma \ref{lemma: CDF transformation}, respectively. Note that this step will not affect the \EOpp problem, since $F^{-1}(\tilde{F}(\cdot))$ is a monotonic transformation. This step might be useful in industrial settings where scores are used in more than one machine learning systems. This line of reasoning is similar to quantile normalization in (bio)statistics \cite{AmaratungaCabreraQuantileNormalization2001,BolstadIrizarryQuantileNormalization03}.

\subsection{Variants of \EOpp algorithm:} \label{app:subsec:variantsEOpp}
These post-processing approaches also allow us to maintain \EOpp between retraining of the models by updating the $F_{c,1}$'s online, as user engagement can be changing dynamically, requiring adjustment to the algorithm \cite{damour20}. In practice, we discretize (for instance, at every percentile or every $10^{-4}$) the CDFs corresponding to the \EOpp transformation and create a linear or higher-order interpolation between the points for applying the transformation. Next, note that the algorithm immediately generalizes to an arbitrary number of characteristics. 
Furthermore, one could relax the strict \EOpp constraint, by considering the following modification to the transformation of the scores:
\begin{align}\label{eq: tradeoff}
\widetilde{s}_{c}(\alpha) = \alpha \times F^{-1}(\tilde{F}(F_{c,1}(s_{c}))) + (1 -  \alpha) \times s_{c},
\end{align}
 where $s_{c}$ denote the score restricted to $C = c$ and $\alpha\in [0, 1]$ is a tuning parameter for linearly mixing the original score with the transformed score. We can tune $\alpha$ to achieve a desirable performance-fairness trade-off (if needed), where larger values of $\alpha$ would bring more fairness (in terms of \textit{EOpp}), possibly at the expense of lower model performance. 

%% file: texfiles/equalOdds.tex
\section{Equalized Odds in Rankings}
\label{sec:EOdds}
Equalized odds (\textit{EOdds}) is a fairness definition which extends equality of opportunity.  In the context of binary classification, \cite{Hardtetal_NIPS2016} defines equalized odds as follows.
\begin{definition}[\EOdds in classification]\label{definition: EOdds} A binary predictor satisfies equalized odds with respect to a (protected) characteristic $C$ and decision $\hat{Y}$, if $\hat{Y}$ is independent of $C$ given $Y$, that is, $\Prob(\hat{Y}=1 \mid C=c_1,Y=y) = \Prob(\hat{Y}=1 \mid C=c_2,Y=y)$
for all $c_1, c_2 \text{ and } y \in \{ 0,1\}$.
\end{definition}
Similarly to the \EOpp definition, the \EOdds definition can be naturally extended to handle the ranking case, where ranks are assigned according to a scoring function $s(X)$.
  
\begin{definition}[\EOdds in rankings]\label{definition: ranking EOdds} A scoring function satisfies equalized odds with respect to a (protected) characteristic $C$ and score $s(X)$ if 
$\Prob(s(X) \leq t \mid C=c_1,Y=y) = \Prob(s(X) \leq t \mid C=c_2,Y=y)$
for all $c_1, c_2, t \text{ and } y \in \{ 0,1\}$. That is, the score distribution is independent of $C$ given the outcome $Y$.
\end{definition}

This is equivalent to the requirement that the distribution of the scores is independent of $C$ given the outcome $Y$. Unlike \EOpp, \EOdds cannot be achieved through a deterministic transformation. \cite{Hardtetal_NIPS2016} provide a randomization mechanism to change binary classification labels in such a way that the derived prediction satisfies \EOpp.  We will now review the binary classification setting, and develop a methodology for the ranking context.  

\subsection{Re-Ranking for Equalized Odds:  }
In the context of classifiers, \EOdds can be achieved by randomizing classifications as follows.  Consider randomly changing the decision of a classification $\hat Y = y$ for group $C = c$ with probability $p_{y,c}$. The resulting randomized classification, satisfies equalized odds whenever the following set of linear constraints are satisfied for all $c_1$, $c_2$ and $y$:
\begin{align*}
    & (1 - p_{y,c_1}) \cdot  \Prob(\hat{Y}=y \mid C=c_1,Y=y)  ~+ \\
    & \qquad \qquad p_{1-y,c_1} \cdot  \Prob(\hat{Y} =1 - y \mid C=c_1,Y=y) \\
 = & (1 - p_{y,c_2}) \cdot  \Prob(\hat{Y}=y \mid C=c_2,Y=y)   ~+ \\
 & \qquad \qquad p_{1-y,c_2} \cdot  \Prob(\hat{Y}=1 - y \mid C=c_2,Y=y).
\end{align*}
  We extend this reasoning to the ranking context. Without loss of generality, assume that the ranking scores $s(\cdot)$ fall in the interval $[0,1)$.  Note that if the scores do not fall in this interval, they can easily be transformed to this interval in a way that does not affect rankings, for instance, by applying an inverse logit transformation. 

Let $I_1,...,I_K$ be a partition of the score domain. That is, choose intervals $I_k = [i_k, i_{k+1})$ for $k = 1,...,K$, where $0 = i_1 < i_2 < \cdots < i_{K+1} = 1$. We will derive a score achieving equalized odds by randomizing the original scores between these intervals. For each category $C=c_i$, define $p_{k,k',c_i}$ (such that $\sum_{k'} p_{k,k',c_i} = 1$ for all $k$ and $c_i$) to be the probability that the score from an item with characteristic $c_i$ is randomly moved from interval indexed by $k$ to interval indexed by $k'$ (for notational compactness, we will write $P$ as the set of these probabilities). Let $Z_{k, c_i}$ be a multinomial random variable with probabilities $(p_{k,1,c_i}, \ldots, p_{k,K,c_i})$. Then, for an item with characteristic $C = c_i$ and score $s(X) \in I_k$, we define $\bar{s}(X;P)$ (or more compactly $\bar{s}(X)$) to be a score chosen in the interval $I_{Z_{k,c_i}}$ according to an arbitrary (for instance, uniform) distribution on $I_{Z_{k, c_i}}$. Note that, 
\begin{align}
\label{eqn:BinDecomp}
   &~\Prob(\bar {s}(X) \in  I_{k'}  \mid  C = c_i, Y = y) \nonumber \\
   = &~\sum\nolimits_k p_{k,k',c_i} \cdot \Prob(s(X) \in I_k \mid C = c_i, Y = y). 
\end{align}

Consequently, the randomized score $\bar {s}(X)$ satisfies equalized odds whenever $p_{k,k',c_i}$'s are chosen to satisfy 
$\Prob(\bar {s}(X) \in I_{k'} \mid  C = c_1, Y = y) = \Prob(\bar {s}(X) \in I_{k'} \mid C = c_2, Y = y)$
for all $k'$ and $y \in \left\{ 0,1 \right\}$.  It is readily seen from Equation \eqref{eqn:BinDecomp} that this specifies a system of linear equations in the interval transition probabilities $P = \{p_{k,k',c_i}\}$. Not only does a solution to this system always exist, but there are infinitely many solutions. Ideally, we would like to use a solution which gives good model performance. To identify a ``best'' solution, let $\phi (F_{X,Y,C}, P)$ be a functional of the data generating distribution $F_{X,Y,C}(\cdot)$ and the interval transition probabilities $P$ such that $\phi (F_{X,Y,C}, P)$ captures some aspect of predictive performance such as mean squared error or area under the receiver operating characteristic curve (ROC-AUC). Note that $F_{X,Y,C}(\cdot)$ is generally not known, but is easily replaced by the empirical version in such situations. Finding the optimal interval transition probabilities can be formulated as a maximization problem 
\begin{align}\label{eqn:EOddsMaximization}
& \max_{P}~~  \phi \left( F_{X,Y,C}, P \right) ~\text{such that}~\nonumber\\
 &\quad \sum\nolimits_k p_{k,k',c_i} \cdot \Prob(s(X) \in I_k \mid C = c_1, Y = y)  \nonumber \\
 = & \quad \sum\nolimits_k p_{k,k',c_i} \cdot \Prob(s(X) \in I_k \mid C = c_2, Y = y)
\end{align}

for all  $c_1$, $c_2, k'$, and $y \in \{0,1\}$. 
\begin{theorem}\label{thm:EOddsValidity}
The randomized score $\bar{s} (X)$ derived from $s(X)$ using interval transition probabilities found as the solution to the optimization problem \eqref{eqn:EOddsMaximization} satisfies
\begin{align*}
\Prob&(\bar{s}(X) \leq t \mid C=c_1,Y=y) = \Prob(\bar{s}(X) \leq t \mid C=c_2,Y=y), 
\end{align*}
for all $c_1$, $c_2$, $y$, and $t$.
\end{theorem}

While the choice of the objective function $\phi$ can be arbitrary, we give two suggestions. One choice is to define $\phi$ as $\Exp{|s(X) - \bar{s}(X)|}$, which gives the least possible average movement of the scores, which were originally chosen to give good performance. The empirical version of this condition can easily be written as a linear equation, which gives the computational convenience that the optimization problem \eqref{eqn:EOddsMaximization} becomes a linear program (LP). Another choice is to maximize for ROC-AUC which boils down to solving a quadratic program (QP) (See Appendix for details).

Let $P^* = \{p_{k, k', c_i}^*\}$ denote the solution of \eqref{eqn:EOddsMaximization}. Then, to apply the \EOdds transformation to each score $s(X) \in I_k$ corresponding to an item in $c_i$, we just need to sample from a multinomial distribution with probabilities $(p_{k,1,c_i}^*, \ldots, p_{k,K,c_i}^*)$ to choose a destination bin index and sample from a, for instance, uniform distribution on the chosen bin. This shows the scalability of the proposed solution. 

\textbf{Time complexity: } In practice, the EOpp/EOdds transformation works by observing the initial score, finding its score bucket, and allocating it to a different score bucket based on a pre-computed mapping for EOpp and based on a pre-computed probability vector for EOdds. Finding the score bucket corresponding to the initial score can be done in $log(K)$ (or even constant) time, where $K$ is the number of score buckets. Allocating it to a different score bucket can be done in O(1) time using HashMaps for EOdds (trivially for EOpp). As the number of items $n$ increases, the new final list can be obtained in $O(n log(K))$ time. 

\textbf{Choice of bins: } Note that the choice of bins are arbitrary but obvious choices include equispaced bins or quantile binning. We recommend choosing intervals based on quantiles for EOpp. However, we experimented with equispaced bins for the EOdds transformation in our social network application and still got a good performance. For EOdds, we note that finer binnings typically result in less degredation to model performance, however in our empirical evaluations (including a social network application), we saw little benefit from using more than 100 equispaced bins or from using quantiles instead of equispaced bins. Given modern linear program solvers can easily handle thousands of bins, we encourage practitioners to use as fine a binning as practical subject to suitable model performance.  

\subsection{Extensions of Equalized Odds: }
\EOdds has several natural extensions such as to general categorical outcomes that are easily handled by adjustments to the constraints in the optimization based methodology for re-ranking.  
Suppose that items are ranked according to a score $s(X)$, and the rankings result in an outcome $y \in {1,...,M}$. For example, in a news feed context, users of a site may engage with articles in multiple ways captured by actions such as ``like,'' ``comment,'' or ``share.'' We extend the definition of equalized odds to ensure that the rankings are fair with respect to any of these outcomes across all characteristics.  

\begin{definition}[Multi-Outcome \EOdds in Rankings]\label{definition: ranking EOdd Multi} A score based ranker satisfies equalized odds with respect to (protected) characteristic $C$ and score $s(X)$ if 
\begin{align*}
\Prob(s(X) \leq t &\mid C=c_1,Y=y)= \Prob(s(X) \leq t \mid C=c_2,Y=y), 
\end{align*}
for all $c_1$, $c_2$, $y \in \left\{ 1,...,M \right\}$ and $t$.
\end{definition}

Achieving multi-outcome \EOdds simply requires extending the linear constraints in optimization problem \eqref{eqn:EOddsMaximization} to all $y \in \left\{ 1,..,M \right\}$. When finding a randomized scoring function as a solution to this modified optimization problem, an analogous result to Theorem \ref{thm:EOddsValidity} holds, but for multi-outcome \textit{EOdds}. 

Another simple modification is to relax the strictness of the equalized odds condition.

\begin{definition}[$\epsilon_0, \epsilon_1$-differentially \EOdds in Rankings]\label{definition: ranking EOdd} A score based ranker satisfies $\epsilon_0, \epsilon_1$-differentially equalized odds with respect to (protected) characteristic $C$ and score $s(X)$ if 
\begin{align*}
\exp{(- \epsilon_y)} \leq \frac{ \Prob(s(X) \leq t \mid C=c_1,Y=y) }{\Prob(s(X) \leq t \mid C=c_2,Y=y)} \leq \exp{(\epsilon_y)} 
\end{align*}
for all $c_1$, $c_2$, $t$ and $y \in \{0,1\}$.
\end{definition}

Ensuring that a randomized ranking score $\bar{s} (X)$ satisfies $\epsilon_0, \epsilon_1$-differentially \EOdds requires conditions that can be expressed as linear constraints, for example through
\begin{align*}
\Prob(\bar{s} (X) \leq t &\mid C=c_1,Y=1) - \exp{(\epsilon)} \Prob(\bar{s}(X) \leq t \mid C=c_2,Y=1) \leq 0
\end{align*}
and similarly for the remaining constraints required. Once again, a simple modification of the constraints in optimization problem \eqref{eqn:EOddsMaximization} leads to a randomized ranking score such that an analogous result to Theorem \ref{thm:EOddsValidity} holds, but for $\epsilon_0, \epsilon_1$-differentially \EOdds. An interesting special case is $\infty, \epsilon_1$-differentially equalized odds, which reduces to $\epsilon_1$-differentially \textit{EOpp}, providing a method to re-rank for an alternative relaxation of \textit{EOpp}. Adjusting the $\epsilon_0, \epsilon_1$ allows the practitioner to strike a desireable balance between fairness and model performance.  

\subsection{Causal Interpretation of Equalized Odds:}\label{subsection: causal interpretation}

We conclude this section by providing an interpretation of \EOdds using causal graphs. A causal graph is a directed acyclic graph (DAG) where the nodes represent variables, and the directed edges in the graph define conditional independence relationships among the variables via the notion of d-separation \cite{Pearl00}. The conditional independence between the scoring variable $s(X)$ (or $S$ in short) and the characteristic $C$ given the observed outcome $Y$ (Definition \ref{definition: ranking EOdds}) can be represented by the absence of paths from $C$ to $S$ that are not d-separated by $Y$ in the underlying causal DAG. We illustrate this through the following examples.

\begin{figure}[!ht]
\nocaption
\centering
\begin{subfigure}[t]{0.3\textwidth}
\includegraphics[width = 0.9\textwidth]{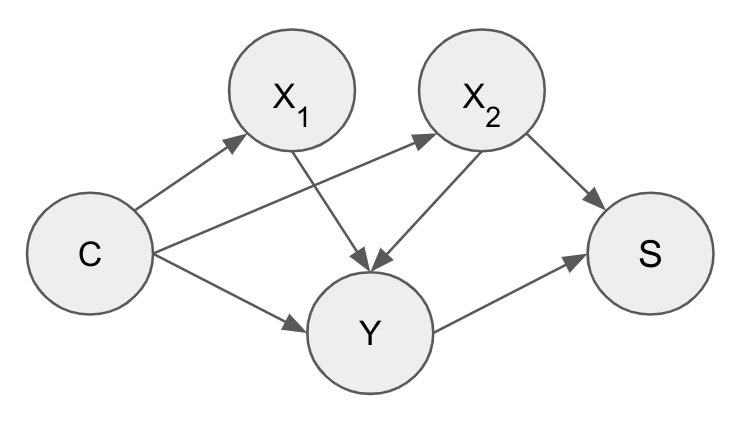}
\caption{\EOdds is not satisfied due to the existence of the directed path $C \to X_2 \to S$.}
\label{fig: EO violation DAG 1}
\end{subfigure}
\hfill
\begin{subfigure}[t]{0.3\textwidth}
\includegraphics[width = 0.9\textwidth]{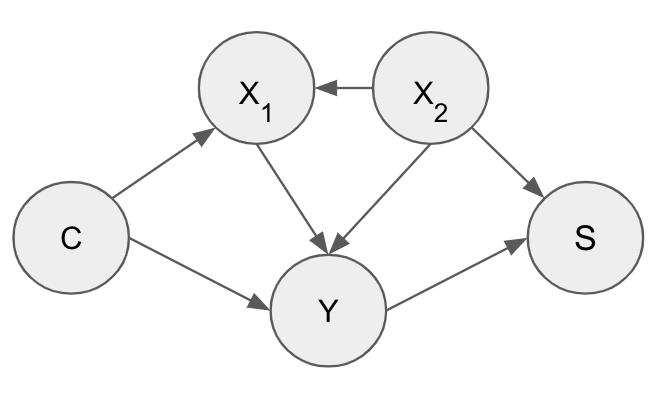}
\caption{\EOdds is not satisfied due to the existence of the collider path $C \to X_1 \leftarrow X_2 \to S$ and the directed edge $X_1 \to Y$.}
\label{fig: EO violation DAG 2}
\end{subfigure}
\hfill
\begin{subfigure}[t]{0.3\textwidth}
\includegraphics[width = 0.9\textwidth]{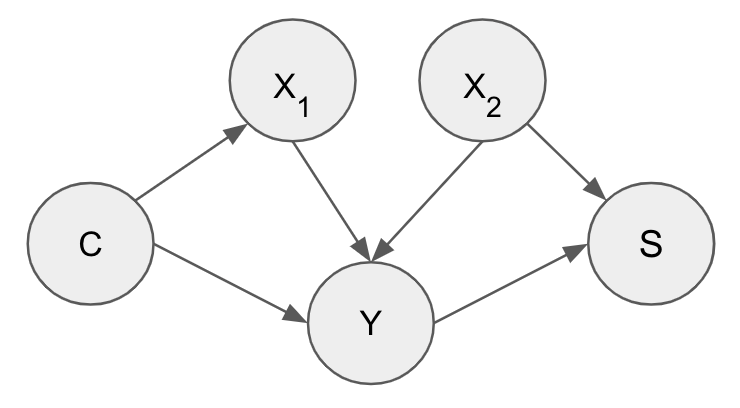}
\caption{\EOdds is satisfied since the only directed path from $C$ to $S$ is d-separated by $Y$.}
\label{fig: EO DAG}
\end{subfigure}
\end{figure}

For simplicity, we assume that no variable has a causal influence on $C$ and $S$ has no causal influence on any variable. These assumptions are reasonable in most situations where $C$ is an exogenous variable representing gender, race, or age, and $S$ is generated through machine learning models based on $Y$ and other variables. The causal DAG corresponding to our first example is depicted in Figure \ref{fig: EO violation DAG 1}. In this case, the \EOdds is not satisfied due to the existence of the directed path $C \to X_2 \to S$ that does not go through $Y$. In other words, the \EOdds condition prohibits the characteristic $C$ from having a causal influence on the score $S$ except through the observed outcome $Y$. In the DAG given in Figure \ref{fig: EO violation DAG 2}, the path $C \to X_1 \leftarrow X_2 \to S$ from $C$ to $S$ is not d-separated by $Y$ since $Y$ is a child of the collider node $X_1$. In other words, the \EOdds condition prohibits a conditional dependency between the characteristic $C$ and the score $S$ given $Y$ through "selection bias" \cite{Pearl09, FelixWinship14}. Finally, the DAG in Figure \ref{fig: EO DAG} satisfies \EOdds since $Y$ is a non-collider node on the only path $C \to Y \to S$ from $C$ to $S$, and hence $C$ and $S$ are d-separated by $Y$.


%% file: texfiles/positionBias.tex
\section{Position Bias Adjustment}\label{sec:positionbias}
To understand the effect of position bias in achieving \EOpp and \textit{EOdds}, consider a recommendation system where for each query $q$, a candidate set of $J$ items $\{d^{(q)}_1, \ldots, d^{(q)}_J\}$ are ranked according to a scoring function $s(X)$ defined on a set of features $X$. The viewer's response to $d^{(q)}_i$ in the recommended list not only depends on the quality of $d^{(q)}_i$ (relative to the viewer) but also depends on the position of $d^{(q)}_i$ in the list. The position bias refers to the fact that the chance of observing a positive response (e.g., \ a click) on an item appearing at a higher position (where the highest position is Position 1) is higher than the chance of observing the same in a lower position. We emphasize that both observed and unobserved items are considered for defining position bias. Consequently, our definition of position bias covers all biases that are due to the position at which an item is ranked, including selection bias \cite{HarrieMaarten20, OvaisiEtAl20} and trust bias \cite{AgarwalEtAl19, VardasbiEtAl20}. Selection bias (users only observing the items ranked in the top k positions), and trust bias (having a higher likelihood to click a higher ranked item among the observed items) both stem from the position at which an item is ranked.

To define \EOpp or \EOdds in the presence of the position bias, we need to take into account the dependency of the response variable $Y$ on the position where the item is shown. To this end, we denote the counterfactual response of an item appearing at position $j$ by $Y(j)$. Furthermore, we denote by $\gamma$ the position of an item in the ranking generated by $s(X)$. Therefore, the observed response is given by $Y(\gamma)$.

\begin{definition}\label{definition: EO with position bias}
A scoring function $s(X)$ of a recommendation system satisfies \EOpp or \EOdds with respect to a characteristic $C$ if
\begin{align}\label{eq: EO with position bias}
\Prob(s(X) \leq &t \mid C=c_1,Y(\gamma)=y) = \Prob(s(X) \leq t \mid C=c_2,Y(\gamma)=y),
\end{align}
for all $t, c_1, c_2$, and for $y = 1$ for \EOpp and $y\in \{0, 1\}$ for \textit{EOdds}.  
\end{definition}

Note that the post-processing approaches discussed in the previous section work under the assumption that $Y(j)$'s are identical for all $j$. To see this, let $\tilde{s}(X)$ denote the score of an item after applying the \EOpp or \EOdds post-processing algorithm and let $\tilde{\gamma}$ denote the corresponding position of the item in the ranking generated by $\tilde{s}(X)$. Then $(\tilde{s}(X), \gamma)$ would satisfy \eqref{eq: EO with position bias} for $\gamma$ equals the position of the item in the ranking without post-processing. However, this does not guarantee that $(\tilde{s}(X), \tilde{\gamma})$ would satisfy  \eqref{eq: EO with position bias} unless $Y(j)$'s are identical for all $j$. Below, we propose adjustments to the \EOpp and \EOdds algorithms for satisfying \eqref{eq: EO with position bias} under the following assumptions on the position bias:

\begin{assumption}\label{assumption: position bias}
For all queries $q$ and all candidate items $d^{(q)}_i$ and all $j > 1$, we assume
\vspace{-0.05in}
\begin{enumerate}
\item Homogeneity: 
$\Prob\big(Y(j) = 1 \mid Y(1) = 1,~ d^{(q)}_i\big) = \Prob(Y(j) = 1 \mid Y(1) = 1)$, and
\item Preservation of Hierarchy: $\Prob \big(Y(j) = 1 \mid Y(1) = 0,~ d^{(q)}_i\big) = 0.$
\end{enumerate}
\end{assumption}
The first assumption states that the position bias is homogeneous over all queries and candidate items. This is a common assumption in the literature \cite{joachims2018, SinghJoachims2019, positionBiasEstimationGoogle2018}. The second assumption states that if a candidate item in a given position does not get a positive response, it cannot get a positive response in a lower position (which is reasonable in most practical settings). Under Assumption \ref{assumption: position bias}, we formally define the position bias as the following positive response decay factor:
\begin{align}
\label{eq: decay factor}
w_j := \Prob(Y(j) = 1 \mid Y(1) = 1)~\text{for $j=1\ldots,J$.}
\end{align}

%

\subsection{Tackling Position Bias for Equality of Opportunity: }
We achieve Eq.\ \eqref{eq: EO with position bias} for \EOpp by learning the conditional CDFs of weighted scores (c.f.\ Lemma \ref{lemma: CDF transformation}) where the weights are given by the inverse of the decay factor defined in Eq.\ \eqref{eq: decay factor}.

\begin{theorem}\label{theorem: weighted CDF}
Let $w_j$ be as in Equation \eqref{eq: decay factor} and let $F_{c, 1}^*(\cdot)$ denote the CDF of the conditional scores $s(X)$ given $Y(\gamma) = 1$ and $C=c$ with weights $1 / w_{\gamma}$. Under Assumption \ref{assumption: position bias}, the transformed scores $\tilde{s}(X) := \sum_{c} F_{c, 1}^*(s(X)) 1_{\{C = c\}}$ satisfy
\begin{enumerate}
\item $\Prob(\tilde\gamma = j \mid C=c_1,~Y(1)=1) = \Prob(\tilde\gamma = j \mid C=c_2,~Y(1)=1)$, and
\item $\Prob(\tilde{s}(X) \leq t \mid C=c_1,~Y(\tilde\gamma)=1) = \Prob(\tilde{s}(X) \leq t \mid C=c_2,~Y(\tilde\gamma)=1),$
\end{enumerate}
for all $j$, $t$ and $c_1, c_2$,  where $\tilde\gamma$ denotes the position of an item based on $\tilde{s}(X)$.
\end{theorem}

The first part of Theorem \ref{theorem: weighted CDF} shows that the weighted CDF transformations guarantee ``fairness of exposure'' with respect to the items with a (counterfactual) positive response at position 1. This is also related to the notion of group fairness parity in \cite{SinghJoachims2019}.

\begin{corollary}\label{corollary: weighted CDF}
For a characteristic $C=c$, let $M_{c} := \Prob(Y(1) = 1 \mid C = c)$ and $v_{obs}(c) := \Prob(Y(\tilde{\gamma}) = 1 \mid C = c)$ be the average merit and the observed exposure (with respect to the scoring function $\tilde{s}(X)$ defined in Theorem \ref{theorem: weighted CDF}), respectively. Then under Assumption \ref{assumption: position bias}, $\tilde{s}(X)$ achieves the group fairness parity, given by the following constraint: $v_{obs}(c_1) / M_{c_1} = v_{obs}(c_2)/M_{c_2}, ~\text{for all}~ c_1, c_2.$
 \end{corollary}

\subsection{Tackling Position Bias for Equalized Odds: }
We achieve Equation \eqref{eq: EO with position bias} for \EOdds by carefully adjusting the counts of positive and negative labels in certain data segments defined by $\{C=c, Y(\gamma) = y, s(X) \in I_k, \gamma = j\}$. To motivate our approach, we first show that if we had access to the counterfactual label $Y(1)$, then the method defined in the previous section would have worked. Then we describe how we can estimate $\Prob(\tilde{s}(X) \in I_k \mid C = c, Y(1)=y)$ by adjusting for the position bias.

\begin{theorem}\label{theorem: EOdds with position bias}
Under Assumption \ref{assumption: position bias}, let $\tilde{s}(X)$ be such that
$\Prob(\tilde{s}(X) \leq t \mid C = c_1, Y(1)=y) = \Prob(\tilde{s}(X) \leq t \mid C = c_2, Y(1)=y),$
for $k \in \{1,\ldots,K\}$, $y \in \{0, 1\}$ and for all $c_1,c_2$. Then $\tilde{s}(X)$ satisfies the equalized odds conditions given in Definition \ref{definition: EO with position bias}.
\end{theorem}

To use Theorem \ref{theorem: EOdds with position bias} in conjunction with the optimization given in \eqref{eqn:EOddsMaximization}, we need to estimate the counterfactual probabilities $\Prob(s(X) \in I_k \mid C = c, Y(1)=y)$ for all $c$, $y$ and $k$. To this end, we define the adjusted positive and negative label counts at position $j$ as follows.
\[
n_{c,1,k}^{(j)\prime} = n_{c,1,k}^{(j)} / w_j ~\text{and}~ n_{c,0,k}^{(j)\prime} = (n_{c,0,k}^{(j)} + n_{c,1,k}^{(j)}) - n_{c,1,k}^{(j)\prime},
\]
where $n_{c,y,k}^{(j)} := \sum_r 1_{\{C_r=c,~ Y_r(\gamma_r) = y,~ s(X_r) \in I_k,~ \gamma_r = j\}}$ is the observed count and $w_j$ is the positive response decay factor defined in \eqref{eq: decay factor}. Then we estimate $\Prob(s(X) \in I_k \mid C = c, Y(1)=y)$ as
\vspace{-0.1in}
\begin{equation}\label{eqn:posBiasBinProbs}
\hat{\Prob}(s(X) \in I_k \mid C = c, Y(1)=y) = \frac{\left(\sum\nolimits_{j} n_{c,y,k}^{(j)\prime}\right)}{ \left(\sum\nolimits_{j,k} n_{c,y,k}^{(j)\prime}\right)}.
\end{equation}
We prove the correctness of this adjustment in Theorem \ref{theorem: consistent estimation}.



\begin{theorem}\label{theorem: consistent estimation}
Under Assumption \ref{assumption: position bias}, $\hat{\Prob}(s(X) \in I_k \mid C = c, Y(1)=y)$ converges to $\Prob(s(X) \in I_k \mid C = c, Y(1)=y)$ almost surely for all $k,c,y$.
\end{theorem}

\subsection{Position Bias Estimation: }
\label{subsec:positionbiasestimation}
The weighted CDF transformations defined in Theorem \ref{theorem: weighted CDF} and the estimation of the counterfactual probabilities $\Prob(s(X) \in I_k \mid C = c, Y(1)=y)$ via \eqref{eqn:posBiasBinProbs} require the positive response decay factors $w_j$'s defined in \eqref{eq: decay factor} to be known or estimated. To estimate $w_j$, we may collect data by randomizing slots and estimate $w_j$ as the ratio of the number of positive responses at position $j$ and position $1$, i.e. 
\begin{align}\label{eq: weight estimation randomized data}
\hat{w}_j = \frac{(\sum_r 1_{\{Y_r(\gamma) = 1,~\gamma_r = j\}}) / (\sum_r 1_{\{\gamma_r = j\}})}{(\sum_r 1_{\{Y_r(\gamma) = 1,~\gamma_r = 1\}}) / (\sum_r 1_{\{\gamma_r = 1\}})}.
\end{align}
Without randomization, we will end up underestimating $w_j$ by using Equation \eqref{eq: weight estimation randomized data} as the items served at position $j$ are expected to be of lower quality than the items served at position 1. More precisely, in the observational data where the items are ranked according to $s(X)$, the conditional distribution of $s(X)$ given $\gamma = 1$ is expected to be stochastically larger than that of $s(X)$ given $\gamma = j$. However, randomly shuffling all items can undesirably harm viewers' experience. A less harmful alternative is to shuffle pairs of items randomly \cite{JoachimsUnbiasedLearningToRankFromBiasedFeedback, positionBiasEstimationGoogle2018}. To estimate the weights from observational data, \cite{positionBiasEstimationGoogle2018} applied the EM algorithm with a parametric click model. Below, we propose a non-parametric approach to estimate the weights from observational data based on importance sampling. To this end, we first estimate the response bias at position $j$ relative to position $j-1$ by correcting for the discrepancy in the distribution of scores in those positions with importance weighting as follows. 

Let $f_j(\cdot)$ denote the conditional density of the observed score at position $j$. For $j\geq 2$, define
\begin{align}\label{eq: eta j}
  \eta_j = \frac{\Exp{Y(\gamma) \frac{f_{j-1}(s(X))}{f_j(s(X))} \mid \gamma = j}} {\Exp{Y(\gamma) \mid \gamma = j-1}}.  
\end{align}

It is straightforward to estimate $\eta_j$ by replacing the conditional density and the conditional expectations with the corresponding empirical estimates. We denote this estimator by $\hat{\eta}_j$. Finally, we estimate $w_j$ by $\hat{w}_j = \prod_{r=2}^{j} \hat{\eta}_r$. We establish the correctness of this method in the following theorem.
 \begin{theorem}\label{theorem: eta j}
 Under Assumption \ref{assumption: position bias}, $\eta_j$ defined in \eqref{eq: eta j} equals $\Prob(Y(j) = 1 \mid Y(1) = 1) ~/~\Prob(Y(j-1) = 1 \mid Y(1) = 1)$ and hence, $w_j$ defined in \eqref{eq: decay factor} equals $\prod_{r = 2}^j \eta_r$.
 \end{theorem}
 
The reason for not estimating the $w_j$ directly by using the importance weights $f_1(s(X))/f_j(s(X))$ is that the distribution of scores at position $1$ might be much different from its counterpart at position $j$, even for not-so-large large $j$. Our adjacent-pairwise importance sampling approach tends to have a lower variance than the direct importance sampling approach since the score distributions of the adjacent positions are expected to be relatively close to each other. However, for practical purposes we recommend to use a truncated version $\hat{w}_j = \prod_{r=2}^{\min(j, T)} \hat{\eta}_r$ for some threshold $T$. This is equivalent to assuming $\eta_j = 1$ for all $j > T$, which is a reasonable practical assumption for most recommendation systems. The overall algorithms for position adjusted \EOpp and \EOdds are given below.


\subsection{Position Bias Adjusted Algorithms}
\begin{algorithm}[!ht]
	\caption{Position Bias Adjusted Equality of Opportunity}\label{algo:EOpp}
	\begin{algorithmic}[1]
        \State \text{\bf{Reranker Training}}
	\State \text{Input: } Score, position, label and characteristic data $(s_i, \gamma_i, y_i, c_i)$, $i = 1,...,n$
	\State Compute the weighted empirical CDF $\hat{F}_{c, 1}^*$ of the conditional scores $s(X)$ given $Y(\gamma) = 1$ and $C=c$ with weights $1 / \hat w_{\gamma}$ computed as in Section ``Position Bias Adjustment'' 
	\State \text{Output: } The empirical distribution functions $\hat{F}_{c, 1}^*$
	\State \text{\bf Reranker Scoring}
	\State \text{Input: } Score $S$, characteristic $C$
	\State {Output: } Fair score, $\tilde S =  \sum_{c} \hat{F}_{c, 1}^*(S) 1_{\{C = c\}}$
	\end{algorithmic}
\end{algorithm}

\begin{algorithm}[!ht]
	\caption{Position Bias Adjusted Equalized Odds}\label{algo:EOdds}
	\begin{algorithmic}[1]
	\State \text{\bf{Reranker Training}}
	\State \text{Input: } Score, position, label and characteristic data $(s_i, \gamma_i, y_i, c_i)$, $i = 1,...,n$ and a partition of the score space $I_1,...,I_K$
	\State Estimate empirical conditional bin probabilities as in \eqref{eqn:posBiasBinProbs} with weights estimated as in Section ``Position Bias Adjustment'' 
	\State Find $P$, the solution to Optimization Problem \eqref{eqn:EOddsMaximization} 
	\State \text{Output: } Vector of interval transition probabilities, $P$
	\State \text{\bf Reranker Scoring}
	\State \text{Input: } Score $S$, characteristic $C$, a partition of the score space $I_1,...,I_K$, interval transition probabilities $P = \{p_{k,k',c}\}$ and distribution functions $F_1,\ldots,F_K$
	\State Determine $k$ such that $S \in I_k$
	\State Choose $k'$ from a multinomial distribution with probabilities $\{p_{k,1,C},\ldots,p_{k,K,C}\}$
	\State Randomly select a point $\tilde S$ within $I_k$ according to $F_k(\cdot)$
	\State {Output: } Fair score, $\tilde S$
	\end{algorithmic}
\end{algorithm}



%% file: texfiles/experiments.tex
\section{Empirical Evaluation}
\label{sec:empirical}
We do a thorough simulation study to validate the need and the efficacy of our algorithms to achieve fairness in web-scale recommender systems. Note that this validation cannot be done with a publicly available dataset since the labels in a validation data must be generated after reranking items with our post-processing algorithms in order to capture the effect of the position bias. Next, we demonstrate the scalability of our methods with an application on a connection recommendation system in a real-life social network platform.

The simulation and empirical studies shown here are to demonstrate that the proposed methods exactly satisfy the stated definitions of fairness (namely \EOpp and \EOdds), with limited impact on (and potentially even improvement to) model performance. Existing work in fairness for recommendation systems considers fairness definitions other than \EOpp and \EOdds and we cannot make meaningful comparisons between fairness definitions, nor can we interpret differences in model performance changes (e.g. stricter notions of fairness often yield larger impact on model predictive performance). For this reason, we omit comparisons with methods of achieving differing notions of fairness and leave the choice of the appropriate fairness criteria to the domain expertise of practitioners who can best assess the suitability of various fairness criteria to the problem at hand. Please see the discussion section below for more on the choice of fairness metrics.

\subsection{Simulation Study:} 
We generate a population of $p = 50000$ items, where each item consists of id $i$, characteristic $C_i \sim \{0, 1\}$, $Y_i(1) \in \{0, 1\}$ and relevance $R_i$. We independently generate $C_i$'s from a $\bern(0.6)$ distribution. The conditional distribution $Y_i(1)$ given $C_i = 0$ is $\bern(0.4)$, and the conditional distribution $Y_i(1)$ given $C_i = 1$ is $\bern(0.5)$. Finally, $R_i \mid (C_i, Y_i(1))$ is generated from
$$\mathcal{N}(0.6Y_i(1) + 2C_i,~ 0.5) + (1 - C_i) \unif[0,~ (1 + Y_i(1))].$$ We consider a recommendation system with $K = 50$ slots. For each query, we randomly select $50$ items from the population and assign a score $s_i = R_i + \mathcal{N}(0, 0.1)$ to each selected item $i \in \{1,\ldots, 50000\}$. The selected items are then ranked according to $s_i$ (in a descending order) and assigned position according to $\textrm{rank}(i)$. Finally, the item at position $j$ gets observed response $Y(j) = Y(1) \times \bern(w_j)$ with position bias $w_j = 1 / \log_2(1+j)$. 
We generate a training dataset based on $100000$ queries and a validation dataset based on $50000$ queries.

\textit{Results: }
We use the adjacent-pairwise importance sampling approach with the threshold $T=30$ for the position bias estimation based on the training dataset. Next, we learn the weighted empirical CDF for \EOpp based on the training data. To learn the position bias-adjusted \EOdds re-ranker based on the training data, we apply an inverse-logit transformation to the score, discretize the score using 100 equally spaced intervals and solve the optimization problem given in \eqref{eqn:EOddsMaximization} with $\Exp{|s(X) - \bar{s}(X)|}$ as the objective function.

\begin{figure*}[!ht]
\includegraphics[width=\textwidth]{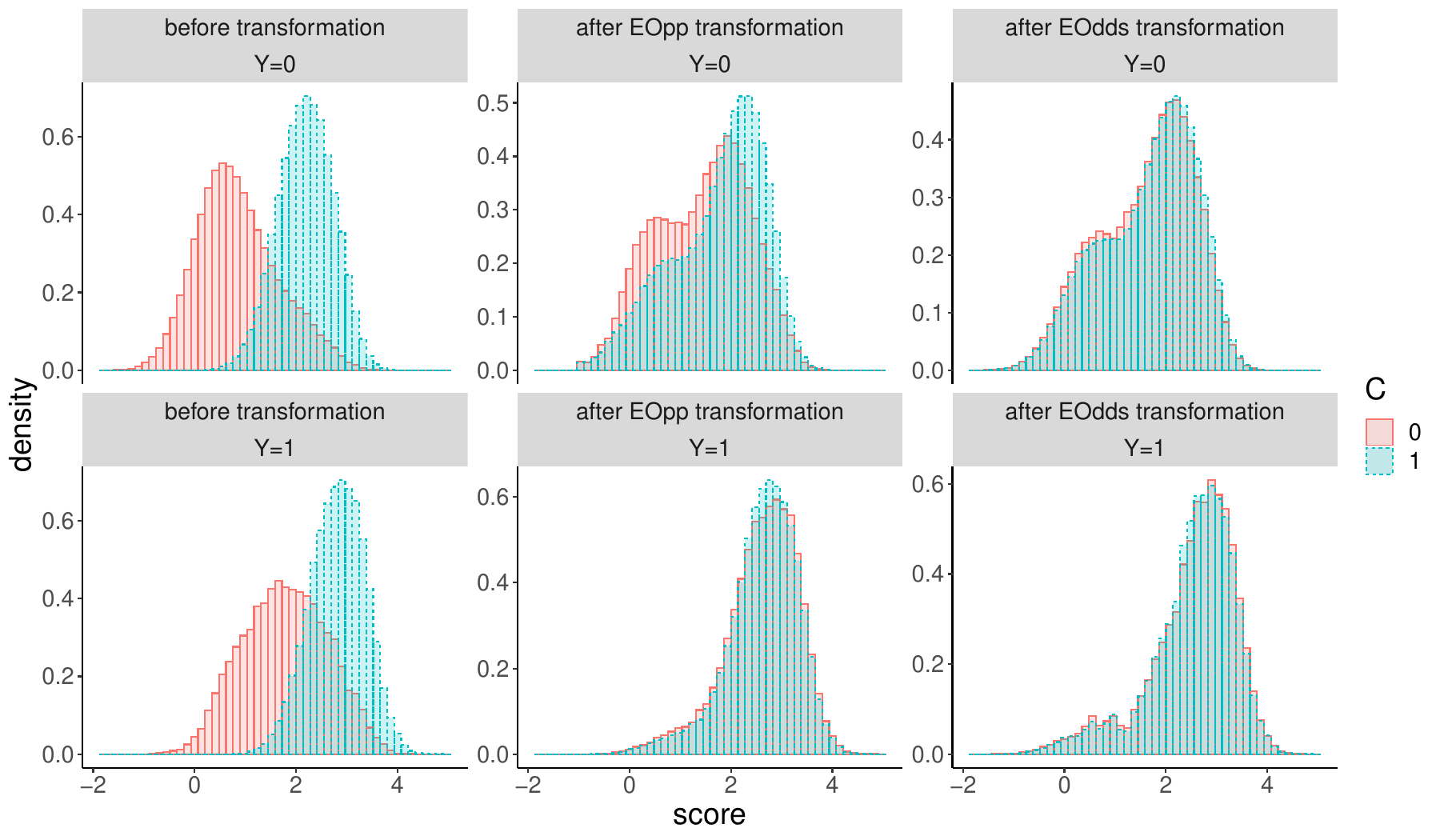}
\caption{
Score distributions before and after \EOpp and \EOdds transformations.
}
\label{fig: EO results}
\end{figure*}
We apply these transformations on the validation data scores and each time regenerate the online labels (with position bias) based on the rankings of the items given by the transformed scores. Figure \ref{fig: EO results} validates the usefulness of our algorithms in achieving \EOpp and \EOdds. Prior to these transformations, it is seen that the conditional score distributions differ greatly between the characteristics.  Post-transforming, the conditional score distributions are identical, as required for equalized odds. Recall that the \EOpp transformation only guarantees to produce identical score distributions across all groups for positive labels, while \EOdds guarantees the same for positive labels as well as for negative labels. These are reflected in Figure \ref{fig: EO results}. 

We implemented the Algorithms in \texttt{R}. Based on the training data with 5 million samples (100k queries with 50 slots), the position bias estimation took 7 seconds, the \EOpp learning took 9 seconds, and the \EOdds learning with 100 bins took 100 seconds on a \texttt{Macbook Pro} with \texttt{3.5 GHz Dual-Core Intel Core i7} processor and \texttt{16 GB 2133 MHz LPDDR3} memory, demonstrating the scalability of the proposed algorithms.

\begin{figure*}[!ht]
 \includegraphics[width=0.7\textwidth]{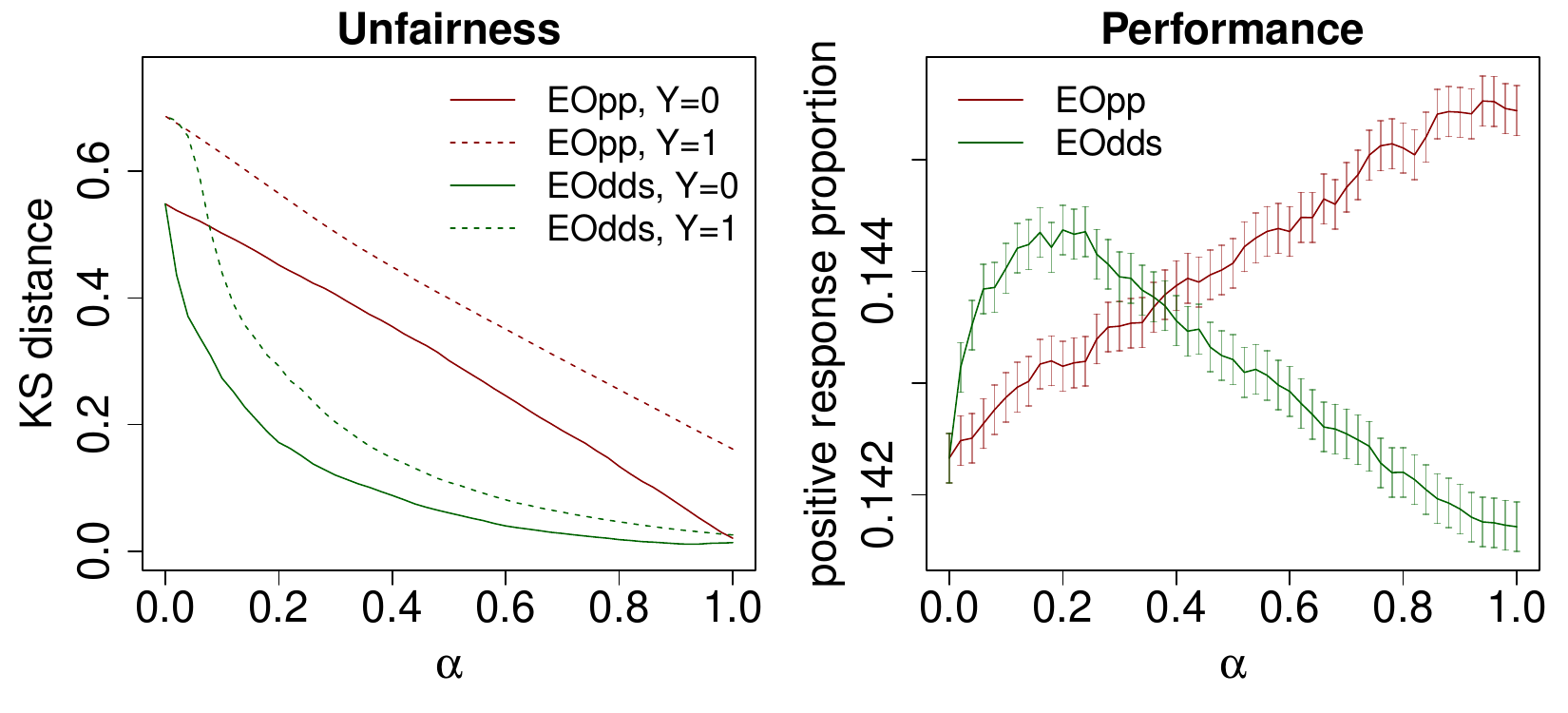}
  \caption{ Unfairness and performance for different $\alpha$. \label{fig: tradeoff}
 }
\end{figure*}

Finally, Figure \ref{fig: tradeoff} demonstrates how a desirable performance-fairness trade-off can be achieved via a linear combination of the transformed scores and the original scores given by,
 $$\alpha \times (\textrm{transformed~score}) + (1 - \alpha) \times (\textrm{original score}),$$ where $\alpha \in [0, 1]$ is a tuning parameter (see \eqref{eq: tradeoff}). Unfairness and performance are measured via Kolmogorov-Smirnov distance (measuring the closeness of the conditional score distributions between groups) and proportion of positive responses respectively. Note that \EOpp and \EOdds are both specified through equating conditional distributions and hence a distribution distance metric (such as the Kolmogorov-Smirnov distance) is a natural choice for measuring ``unfairness'' (or deviations from the desired fairness condition). The performance measured through the proportion of positive labels (cf. click-through-rate in online recommendations) and the error bars in the figure represent one standard deviation of uncertainty. Here, we observe the following:
 \begin{enumerate}
 \item The unfairness decreases to zero as $\alpha$ increases, except for \EOpp corresponding to negative labels (as expected).
 \item The unfairness corresponding to the \EOpp transformation is strictly lower for all values of $\alpha$, while the performance of \EOpp is strictly better for $\alpha \geq 0.4$.
 \item The performance corresponding to the \EOpp transformation is monotonically increasing with $\alpha$. This serves as a counterexample to the popular belief that fairness and performance are always conflicting properties of recommender systems. The improvement in performance by enforcing fairness has also been observed in some recent works \cite{BelloHonorio20, IslamEtAl21, MaityEtAl20}. We note that this is possible, for example, when a more relevant group of candidates gets underexposed through a recommendation system (and we can achieve a correct exposure through unfairness mitigation).
 \end{enumerate}
 

%
%
%
%
%

%% file: texfiles/application.tex
\subsection{Social Network Application: }
Friend or connection recommendation systems are used by many large social network companies such as Facebook, Instagram, LinkedIn, etc. These systems suggests members to connect with others, in order to build their network. Here, a member sending an invitation to connect with a suggested candidate 
can be viewed as a positive outcome. We apply our methodology to (proprietary) training data used for such a product containing historical recommendations and labels indicating whether an invitation has been sent. Members are categorized as infrequent members (IMs; members who are less active on the platform) or frequent members (FMs) who tend to have greater rates of engagement and correspondingly higher representation in the training data. In this example there is potential for the model to not only be biased against IMs, but for that bias to be reinforced over time, leading to a system that is optimized for the benefit of members who are already highly engaged on the site (also known as ``popularity bias'' or the ``rich getting richer'' phenomenon \cite{abdollahpouri2019unfairness}). 

We see an opportunity to apply fairness notions, as a debiasing mechanism, to adjust for the exposure of IMs as candidates being recommended, thus giving them opportunities to be shown and invited.  In our experiments, we applied both the \EOpp and \EOdds reranker to give qualified IMs and FMs equal representation in recommendations. We build the required transformations using two weeks of training data. While serving we apply the transformation on the top 100 candidates. 
The serving was done via a discretized CDF (with 10000 points) for \EOpp and through the estimated transition probabilities (based on 100 bins) for \textit{EOdds}. Due to the simple transformation in both approaches, we did not see any drastic gain in latency, which is a core-requirement in large-scale recommender systems.  The results of the A/B tests on real member traffic are shown in Table \ref{tab:pymk}. 
\begin{table}[!ht]
\centering
\begin{tabular}{|l|c|c|c|c|c|c|c|c|} 
\hline
Invitation & \multicolumn{2}{c|}{Equality of Opportunity} & \multicolumn{2}{c|}{Equalized Odds}\\
Metrics & IM & FM & IM & FM\\ 
\hline
Sent & + 5.72\%& Neutral &+ 2.77\%& Neutral \\
Accepted & + 4.85\% & Neutral &+ 2.26\% & Neutral \\
\hline
\end{tabular}
\caption{A/B Experimentation results for rerankers. 
}
\label{tab:pymk}
\end{table}

Two of the cornerstone metrics for evaluating such experiments are invitations sent and invitations accepted. While one might expect that this post-processing approach would shift invitations away from FMs to IMs and may be detrimental to the metrics for FMs, we were heartened to see more invites being sent to and accepted by IMs without any statistically significant negative impact (pvalue > 0.05) on the FMs.
This highlights that re-ranking approaches pursuing fairness have improved overall recommendation quality.

%% file: texfiles/conclusion.tex
\section{Discussion}
\label{sec:discussion}


We have proposed post-processing methods for handling equality of opportunity and equalized odds in relevance score-based rankings and suggested simple mechanisms for controlling the fairness versus performance trade-off. Our post-processing approaches are applicable in many internet-industry applications as they are tailored towards scalability, have a relatively low engineering footprint, and, unlike in-processing methods, can be relatively easily added on top of existing machine-learning/AI pipelines. 

We explicitly handled the position bias issue while implicitly assuming that the labels are unbiased. In the presence of label bias, methodology such as \cite{JiangNachum20}, which provides a mechanism for imputing debiased labels, can be applied prior to the methods developed in this paper. We considered equality of opportunity and equalized odds as the definition of fairness, which amounts to providing equal opportunity or exposure to equally qualified candidates in a ranking irrespective of their protected attribute status. Here equal qualification is determined by the observed response (i.e., $Y=1$ corresponds to qualified candidates) after correcting for position bias. We note that there are several other existing definitions of fairness that we do not cover in this work, including predictive rate parity \cite{ZafarEtAl17}, demographic parity \cite{DworkEtAl12}, counterfactual fairness \cite{KushnerEtAl17}. These fairness criteria are often conflicting \cite{Saravanakumar21}, and the choice of a fairness criterion should be application-specific. 

The relative merits of various fairness definitions has been studied in other works including \cite{Corbett17} and \cite{corbett2018} and we encourage practitioners to familiarize themselves with the possible benefits and pitfalls before deciding on a fairness criteria. Equality of opportunity may be an appropriate choice when balancing exposure of recommendations resulting in a ``positive'' outcome is across groups is desirable and equalized odds may be useful when balancing exposure conditional on any outcome label is important. Works such as \cite{jacobs13} and \cite{Orhan21} demonstrate how well-intentioned fairness initiatives can lead to unintended consequences. Ultimately, we leave it to the practitioners to use their domain expertise to judge the applicability of equality of opportunity or equalized odds and suggest that users carefully monitor the results of fairness mitigation over time to safeguard against any unforeseen harm.

This research can benefit groups that are currently disadvantaged by large-scale automatic decision systems.
While we do not see any clear negative outcomes of this work, we remain mindful that fairness is a dynamic problem, and we have addressed it from a static perspective. This work has also not addressed intersectionality explicitly, though the framework can handle some intersectional questions. Good practice requires monitoring the fairness performance of the tools we have developed and as well as assessing whether they have unintended intersectional consequences.





%% file: texfiles/appendix.tex
\section{Appendix}

\subsection{Equalized Odds in Rankings}

In the main text, we described how we can solve for Equalized Odds when we define $\phi$ as $\Exp{|s(X) - \bar{s}(X)|}$. Here we give the details when we choose $\phi$ as the ROC-AUC. 

A Riemann approximation to the ROC-AUC based on the partition $I_1,...,I_K$ can be computed as 
\begin{align*}
\sum\nolimits_c &\Prob(C = c) \cdot \sum\nolimits_{k=1}^K  \Prob(\bar{s}(X) \in I_k \mid Y = 0)  \sum\nolimits_{k' \geq k} \Prob(\bar{s}(X) \in I_{k'} \mid Y = 1).
\end{align*}
It is readily seen from Equation \eqref{eqn:BinDecomp} that this is a quadratic function of the transition probabilities. Therefore, when using this choice of objective, the optimization problem \eqref{eqn:EOddsMaximization} becomes a quadratic program (QP) with $K^2\cdot L \cdot (L-1)$ variables for a characteristic $C$ with $L$ categories.

\subsection{Position Bias Estimation}
 \begin{figure}[!ht]
 \centering
 \includegraphics[width=0.5\linewidth]{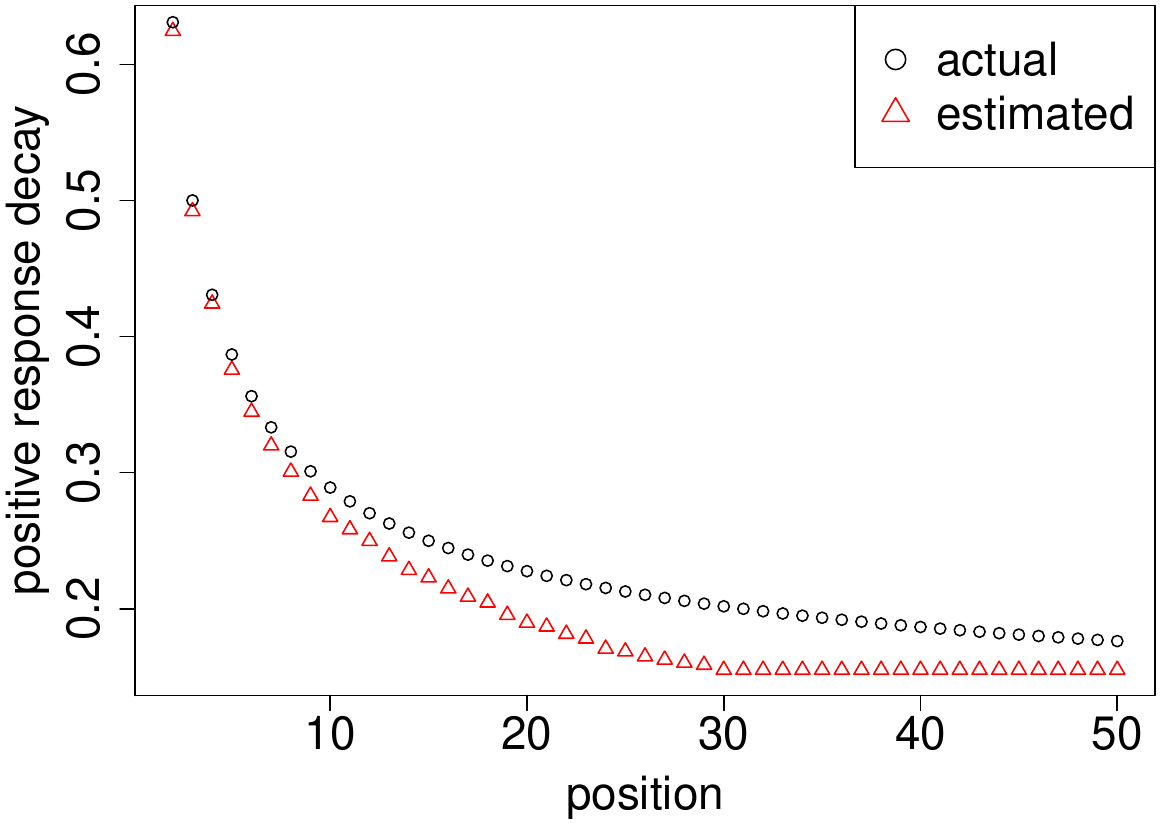}
 \caption{Estimating the position bias $w_j = 1 / \log2(1+j)$ for $j = 2,\ldots, 50$, using the adjacent-pairwise importance sampling approach with the threshold $T=30$. } 
 \label{fig: position bias}
 \end{figure} 
 
  We use the estimator $\hat{w}_j = \prod_{r=2}^{\min(j, T)} \hat{\eta}_r$ with $T = 30$ defined in Section ``Position Bias Adjustment'' 
  to estimate the position bias for each $j = 2,\ldots, 50$. Although we estimate each $\eta_r$ unbiasedly, but the product of the estimators of $\eta_r$’s is no longer guaranteed to be unbiased. This explains the underestimation pattern in Figure \ref{fig: position bias}. We deliberately chose a simulation setting where the estimation of $w_i$ is challenging to demonstrate the insensitivity of our mitigation techniques to the position bias estimation errors.

 \subsection{Proofs}
 \begin{proofof}{Lemma \ref{lemma: CDF transformation}}
 Let $s_{c,1}$ denote the random variable corresponding to the score $s(X)$ restricted to $C = c$ and $Y=1$. Since $F_{c,1}$ is the CDF of $s_{c,1}$, the distribution of $F_{c,1}(s_{c,1})$ is $Uniform[0,1]$ for all $c$. Hence these transformed scores satisfy equality of opportunity across all thresholds. 
 \end{proofof}
 
 \begin{proofof}{Theorem \ref{theorem: weighted CDF}}
 We first prove the results using the following claims and then prove the claims.\\~\\
{\bf Claim 1}: For all $j > 1$ and for all $t$,
\begin{align*}
&~ P(s(X) \leq t \mid \gamma = j,~Y(j) = 1,~ C = c)\\
= &~ P(s(X) \leq t \mid \gamma = j,~ Y(1) = 1,~ C = c).
\end{align*}~\\
{\bf Claim 2}: For all $j > 1$, 
\begin{align*}
P(\gamma = j &\mid Y(\gamma) = 1,~ C = c) = \frac{w_j P(\gamma = j \mid Y(1) = 1,~ C = c)}{\sum_r w_r P(\gamma = r \mid Y(1) = 1,~ C = c)}.
\end{align*}

Using Claims 1 and 2, we show that the CDF of $s(X)$ given $Y(1) = 1$ and $C=c$ equals $F_{c, 1}^*$.

\begin{align*}
&\quad F_{c, 1}^*(t)  \\
&:= ~ \sum_j  \bigg\{ P(s(X) \leq t \mid \gamma = j,~Y(j) = 1,~ C = c)~ \frac{P(\gamma = j \mid Y(\gamma) = 1, C= c) / w_j}{\sum_r P(\gamma = r \mid Y(\gamma) = 1, C= c) / w_r} \bigg\} \nonumber \\
&= ~ \sum_j \bigg\{ P(s(X) \leq t \mid \gamma = j,~Y(1) = 1,~ C = c) ~\frac{P(\gamma = j \mid Y(1) = 1, C= c)}{\sum_r P(\gamma = r \mid Y(1) = 1, C= c)} \bigg\} \nonumber \\
&= ~  \sum_j \big\{ P(s(X) \leq t \mid \gamma = j,~Y(1) = 1,~ C = c) ~ P(\gamma = j \mid Y(1) = 1, C= c) \big\} \nonumber \\
&= ~ \sum_j P(s(X) \leq t,~\gamma = j  \mid Y(1) = 1,~ C = c)\nonumber\\
&= ~P(s(X) \leq t \mid Y(1) = 1,~ C = c).\nonumber
\end{align*}

The first equality follows from the definition of $ F_{c, 1}^*$, we use Claims 1 and 2 in the second equality, the third equality follows from the fact that $\sum_j P(\gamma = j \mid Y(1) = 1, C= c) = 1$, and the fourth equality follows from the fact that 
$$\cup_j \big( \{s(X) \leq t\} \cap \{\gamma = j\} \big) = \{s(X) \leq t\}.$$

We have shown that the conditional distribution of $s(X)$ given $Y(1) = 1$ and $C=c$ equals $ F_{c, 1}^*$. This implies that the conditional distribution of $\tilde{s}(X) := F_{c, 1}^*(s(X))$ given $Y(1) = 1$ and $C=c$ is $Uniform[0,~1]$ for all $c$. Therefore, the conditional distributions of the position $\tilde\gamma$ given $Y(1) = 1$ and $C=c$ are identical for all $c$. To see this, note that
\begin{align}\label{eq: gamma_j distribution}
P(\tilde\gamma \leq j &\mid Y(1) = 1,~ C=c) = P(\tilde{s}(X) > U_{(K - j)} \mid Y(1) = 1,~ C=c),
\end{align}
where $K$ is the total number of positions, $U_{(0)} = 0$ and for $j < K$, $U_{(K - j)}$ is the $(K - j)$-th order statistic corresponding to $K$ i.i.d.\ uniform samples. Note that the right hand side of \eqref{eq: gamma_j distribution} does not depend on $c$ since the conditional distribution of $\tilde{s}(X) := F_{c, 1}^*(s(X))$ given $Y(1) = 1$ and $C=c$ is $Uniform[0,~1]$ for all $c$.  This completes the proof of the first part of the theorem.

To prove the second part, we will use Claims 1 and 2 with $(\tilde{s}(X),~\tilde\gamma)$ instead of $(s(X),~\gamma)$.
\begin{align*}
& \quad P(\tilde{s}(X) \leq t \mid C=c,~Y(\tilde\gamma)=1) \\
&= \sum_j \bigg\{ P(\tilde{s}(X) \leq t \mid \tilde\gamma = j,~Y(j) = 1,~ C = c) ~ P(\tilde\gamma = j \mid Y(\tilde\gamma) = 1, C= c) \bigg\}\\
& = \frac{\sum_j w_j~P(\tilde{s}(X) \leq t,~\tilde\gamma = j  \mid Y(1) = 1,~ C = c)}{\sum_r w_r P(\tilde\gamma = r \mid Y(1) = 1, C= c)}.
\end{align*}
Now note that it follows from the first part of theorem that the denominator is identical for each $c$. Furthermore, the numerator does not depend on $c$, since it follows from \eqref{eq: gamma_j distribution} and the independence of $(\tilde{s}(X) \mid Y(1) = 1)$ and $C$ that
\begin{align*}
&~P(\tilde{s}(X) \leq t,~\tilde\gamma = j  \mid Y(1) = 1,~ C = c) \\
= &~P(U_{(K - j)} < \tilde{s}(X) \leq \min\{U_{(K - j + 1)},~ t\} \mid Y(1) = 1).
\end{align*}
This completes the second part of the theorem.\\~\\
{\bf Proof of Claim 1}: From the second part of Assumption \ref{assumption: position bias}, it follows that
\begin{align*}
& ~ P(s(X) \leq t \mid \gamma = j,~Y(j) = 1,~ C = c) \\
=  & ~ P(s(X) \leq t \mid \gamma = j,~Y(j) = 1,~ Y(1) = 1,~ C = c).
\end{align*}
Therefore, the results follows from the first part of Assumption \ref{assumption: position bias} that ensures that $Y(j)$ is independent of $s(X)$,~ $\gamma$ and $C$ given $Y(1)$.\\~\\
{\bf Proof of Claim 2}: For notational convenience, we denote the conditional probabilities given $C=c$ by $P_c(\cdot)$. Using the second part of Assumption \ref{assumption: position bias} and then applying the Bayes' theorem, we get
\begin{align*}
&~ P_c(\gamma = j \mid Y(\gamma) = 1) \\
=&~ P_c(\gamma = j \mid Y(\gamma) = 1,~ Y(1)=1) \\
=&~ \frac{P_c(Y(\gamma) = 1 \mid Y(1)=1, ~\gamma = j) P_c(\gamma = j \mid Y(1)=1)}{\sum_r P_c(Y(\gamma) = 1 \mid Y(1)=1, ~\gamma = r) P_c(\gamma = r \mid Y(1)=1)} \\
=&~ \frac{P(Y(j) = 1 \mid Y(1)=1) P_c(\gamma = j \mid Y(1)=1)}{\sum_r P(Y(r) = 1 \mid Y(1)=1) P_c(\gamma = r \mid Y(1)=1)}\\
=&~ \frac{w_j P_c(\gamma = j \mid Y(1)=1)}{\sum_r w_r P_c(\gamma = r \mid Y(1)=1)}.
\end{align*}
The seconds last equality follows from the first part of Assumption \ref{assumption: position bias}, and the last equality follows from the definition of $w_j$. 
\end{proofof}

\begin{proofof}{Corollary \ref{corollary: weighted CDF}}
For notational convenience, we denote the conditional probabilities given $C=c$ by $P_c(\cdot)$. Note that
\begin{align*}
\frac{v_{obs}(c)}{M_{c}} &=\frac{\sum_j P_c(Y(j) = 1 \mid Y(1) = 1,~\tilde{\gamma} = j) P_c(Y(1) = 1,~\tilde{\gamma} = j)}{P_c(Y(1) = 1)}\\
&=\frac{\sum_j P(Y(j) = 1 \mid Y(1) = 1) P_c(Y(1) = 1,~\tilde{\gamma} = j)}{P_c(Y(1) = 1)}\\
&=\sum_j w_j P_c(\tilde{\gamma} = j \mid Y(1) = 1),
\end{align*}
where $w_j$ is as in Theorem \ref{theorem: weighted CDF} and the second equality follows from the first part of Assumption \ref{assumption: position bias}. Therefore, the result follows from the first part of Theorem \ref{theorem: weighted CDF}.
\end{proofof}

\begin{proofof}{Theorem \ref{thm:EOddsValidity}}
Fix a $t \in [0,1)$.  Assume $t \in I_k$, i.e. $i_k \leq t < i_{k+1}$. Then, 
\begin{align*}
&~P(\bar {s}(X) \leq t  \mid C = c, Y = y) \\
= &~ P(\bar {s}(X) \leq t \mid \bar {s}(X) \in I_k  , C = c, Y = y) ~\times~ P(\bar {s}(X) \in I_k \mid C = c, Y = y) \\
= &~ F_k(t) ~\times~ P(\bar {s}(X) \in I_k \mid C = c, Y = y)
\end{align*}
Now, $F_k(t)$ does not depend on $y$ or $c$, and the constraints satisfied in the optimization problem imply that 
$P(\bar {s}(X) \in I_k \mid C = c, Y = y)$
does not depend on $c$ conditionally on $y$.  It follows that equalized odds is satisfied in the ranking sense.  
\end{proofof}

\begin{proofof}{Theorem \ref{theorem: EOdds with position bias}}
Note that without loss of generality, we can assume that the $\tilde{s}(X))$ given $Y(1) = 1$ and $C=c$ is $Uniform[0,1]$. This is because we can transform the score using the same CDF function (since they have the same distribution) to make them $Uniform[0,1]$ for all $c$. Then it follows from the arguments given the proof of Theorem \ref{theorem: weighted CDF} that $P(\tilde{s}(X) \leq t \mid C=c,~Y(\tilde\gamma)=1)$ is independent of $c$.
\end{proofof}

\begin{proofof}{Theorem \ref{theorem: consistent estimation}}
For notational convenience, we denote the conditional probabilities given $C=c$ by $P_c(\cdot)$. By applying Bayes' Theorem, we get
 \begin{align*}
 \Prob_c(s(X) \in I_k &\mid Y(1)=y) ~=~ \frac{\Prob_c(Y(1) = y,~ s(X) \in I_k)}{\sum_{\ell} \Prob_c(Y(1) = y,~ s(X) \in I_{\ell})}.
 \end{align*}
Next, note that 
 \begin{align*}
& \qquad \Prob_c(Y(1) = 1,~ s(X) \in I_{\ell}) \\
& = ~ \sum_j \Prob_c(Y(1) = 1,~ \gamma = j,~ s(X) \in I_{\ell}). \\
& =  ~ \sum_j \bigg\{ \Prob_c(s(X) \in I_{\ell} \mid Y(1) = 1,~ \gamma = j)  ~ \times~ \Prob(Y(1) = 1,~ \gamma = j) \bigg\} \\
& =  ~ \sum_j \bigg\{ \Prob_c(s(X) \in I_{\ell} \mid Y(j) = 1,~ \gamma = j) ~\times~ \Prob_c(Y(1) = 1,~ \gamma = j) \bigg\} \\
& =  ~ \sum_j \frac{\Prob_c(s(X) \in I_{\ell}, Y(j) = 1,~ \gamma = j)}{\Prob_c(Y(j) = 1 \mid Y(1) = 1,~ \gamma = j)} \\
& =  ~ \sum_j \frac{\Prob_c(s(X) \in I_{\ell}, Y(j) = 1,~ \gamma = j)}{\Prob(Y(j) = 1 \mid Y(1) = 1)} \\
&  = ~\sum_j  \frac{\Prob_c(s(X) \in I_{\ell}, Y(j) = 1,~ \gamma = j)}{w_j}.
 \end{align*}
 The first, second and the fourth equalities follow from the definition of conditional probability, the third equality follows from Claim 1 in the proof of Theorem \ref{theorem: weighted CDF}, the fifth equality follows from Assumption \ref{assumption: position bias} and the last equality follows from the definition of $w_j$.
 
 Now it follows from the strong law of large numbers that $\frac{n_{c,1,\ell}^{(j)}}{n_c}$ converges almost surely to $\Prob_c(s(X) \in I_{\ell}, Y(j) = 1,~ \gamma = j)$, where $n_c$ is the number of samples corresponding to $C=c$. Hence, we have
 \[
 \hat{\Prob}_c(s(X) \in I_k \mid Y(1)=1) \overset{a.s}{\longrightarrow}  \Prob_c(s(X) \in I_k \mid Y(1)=1),
 \]
 where $\overset{a.s}{\longrightarrow}$ denotes almost sure convergence.
 
 Finally, note that
\begin{align*}
~ &\Prob_c(Y(1) = 0,~ s(X) \in I_{\ell}) \\
=~ &  \Prob_c(s(X) \in I_{\ell}) - \Prob_c(Y(1) = 1,~ s(X) \in I_{\ell}) \\
=~ &  \Prob_c(s(X) \in I_{\ell}) - \sum_j  \frac{\Prob_c(s(X) \in I_{\ell}, Y(j) = 1,~ \gamma = j)}{w_j}.
\end{align*}
 
 Therefore, 
 $\hat{\Prob}_c(s(X) \in I_k \mid Y(1)=0) \overset{a.s}{\longrightarrow}  \Prob_c(s(X) \in I_k \mid Y(1)=0)$
 follows from the strong law of large numbers similarly.
\end{proofof}

\begin{proofof}{Theorem \ref{theorem: eta j}}
Note that
\begin{align*}
& \Exp{Y(\gamma) \frac{f_{j-1}(s(X))}{f_j(s(X))} \mid \gamma = j} \\ 
=~ & \int \Exp{Y(\gamma) \frac{f_{j-1}(z)}{f_j(z))} \mid \gamma = j, s(X) = z} f_{j}(z) dz \\
=~ & \int \Prob(Y(j) = 1 \mid \gamma = j,~ s(X) = z)~ f_{j-1}(z) dz \\
=~ & \int \Prob(Y(j) = 1 \mid Y(1) = 1,~ \gamma = j - 1,~ s(X) = z)  ~\times\\
&\qquad \qquad ~\Prob(Y(1) = 1 \mid \gamma = j,~ s(X) = z)  f_{j-1}(z) dz \\
=~ & \Prob(Y(j) = 1 \mid Y(1) = 1) ~\times~ \int \Prob(Y(1) = 1 \mid  s(X) = z) ~f_{j-1}(z)dz.
\end{align*}

The second last equality follows from the Preservation of Hierarchy assumption in Assumption \ref{assumption: position bias}, and the last equality follows from the Homogeneity assumption in Assumption \ref{assumption: position bias}. Similarly, we have
\begin{align*}
 & ~ E\left[ Y(\gamma) \mid \gamma =  ~ j - 1 \right] \\
 = & ~ P(Y(j-1) = 1 \mid Y(1) = 1)  \times  \int  P(Y(1) = 1 \mid s(X) = z) f_{j-1}(z)dz.
\end{align*}

This completes the proof of the first part of the theorem, and the second part follows directly from the first part as $P(Y(j-1) = 1 \mid Y(1) = 1) = 1$ for $j = 2$. 
\end{proofof}